\documentclass{article}

\usepackage[nonatbib,preprint]{nips_2018}

\usepackage[utf8]{inputenc} % allow utf-8 input
\usepackage[T1]{fontenc}    % use 8-bit T1 fonts
\usepackage{booktabs}       % professional-quality tables
\usepackage{amsfonts}       % blackboard math symbols
\usepackage{nicefrac}       % compact symbols for 1/2, etc.
\usepackage{microtype}      % microtypography
\usepackage{graphicx}
\usepackage{epstopdf}
\usepackage{color}
\usepackage{subcaption}
\usepackage{amsthm}
\usepackage{amsmath}
\usepackage{amssymb}
\usepackage{hyperref}       % hyperlinks
\usepackage{cleveref}
\usepackage{url}            % simple URL typesetting
\usepackage{paralist}
\usepackage{color}
\usepackage{setspace}
\usepackage{algorithm}
\usepackage{algorithmic}
\usepackage{tikz}
\usepackage{url}
\usepackage{caption}
\usepackage{bigstrut}
\usepackage{multirow}
\usepackage{paralist}
\usepackage{array}
\usepackage{tabularx}
\usepackage{mdwmath}
\usepackage{mdwtab}
\usepackage{accents}
\usepackage{bbm}
\usepackage{bm}
\usepackage{xspace}
\usepackage{comment}

\newtheorem{theorem}{Theorem}[section]

\newtheorem{definition}[theorem]{Definition}

\newtheorem{goal}[theorem]{Goal}

\newcommand{\etal}{\emph{et al.}\xspace}

\newcommand{\indicator}[1]{\mathbbm{1}\{#1\}}

\newcommand{\numclasses}{C}
\newcommand{\nummodels}{D}
\newcommand{\numepochs}{n_e}

\newcommand{\ci}{\textsc{CI$(M,\hat\delta,x)$}}
\newcommand{\bt}{\textsc{BT$(M,T)$}}

\newcommand{\clf}[1]{\emph{clf}$_{#1}$}

\newcommand{\rs}{ResNet\xspace}
\newcommand{\rss}{ResNet-110\xspace}
\newcommand{\rsl}{ResNet-50-v2\xspace}

\newcommand{\cf}{\textsc{CIFAR}\xspace}
\newcommand{\cft}{\textsc{CIFAR-10}\xspace}
\newcommand{\cfoh}{\textsc{CIFAR-100}\xspace}
\newcommand{\svhn}{\textsc{SVHN}\xspace}
\newcommand{\inet}{\textsc{Imagenet}\xspace}

\newcommand{\ignore}[1]{}

\DeclareMathOperator*{\argmin}{arg\,min}
\newcommand{\Mysubfloat}[4]{\subfloat[#1]{\includegraphics[width=#2\textwidth]{#3}\label{#4}}}

\DeclareGraphicsExtensions{.eps,.jpg,.png}
\title{Dynamically Sacrificing Accuracy for Reduced Computation: Cascaded Inference Based on Softmax Confidence}

\author{
  Konstantin Berestizshevsky\\
  School of Electrical Engineering\\
  Tel Aviv University\\
  \texttt{konsta9@mail.tau.ac.il} \\
   \And
   Guy Even \\
   School of Electrical Engineering \\
   Tel Aviv University \\
   \texttt{guy@eng.tau.ac.il} \\
}

\begin{document}

\maketitle

\begin{abstract}

  We study the tradeoff between computational effort and classification
  accuracy in a cascade of deep neural networks. During inference, the user
  sets the acceptable accuracy degradation which then automatically
  determines confidence thresholds for the intermediate classifiers. As soon
  as the confidence threshold is met, inference terminates immediately
  without having to compute the output of the complete network. Confidence
  levels are derived directly from the softmax outputs of intermediate
  classifiers, as we do not train special decision functions. We show that
  using a softmax output as a confidence measure in a cascade of deep neural
  networks leads to a reduction of $15\%-50\%$ in the number of MAC
  operations while degrading the classification accuracy by roughly $1\%$.
  Our method can be easily incorporated into pre-trained non-cascaded
  architectures, as we exemplify on \rs. Our main contribution is a method
  that dynamically adjusts the tradeoff between accuracy and computation
  without retraining the model.

\end{abstract}

%%%%%%%%%%%%%%%%%%%%%%%%%%%%%%%%%%%%%%%%%%%%%%%%%%%%%%%%%%%%%%%%%%%%%%%%%%
%%%%%%%%%%%%%%%%%%%%%%%%       Introduction      %%%%%%%%%%%%%%%%%%%%%%%%%
%%%%%%%%%%%%%%%%%%%%%%%%%%%%%%%%%%%%%%%%%%%%%%%%%%%%%%%%%%%%%%%%%%%%%%%%%%

\section{Introduction}

State-of-the-art Deep Neural Networks (DNNs) usually consist of hundreds of
layers and millions of trainable weights. At inference time, this translates
into billions of multiply-accumulate operations (MACs) for a single
input~\cite{sze2017efficient}. The training process of models is a
computationally intensive task that is performed once. After training is
completed, the trained model is used for inference. Inference requires fewer
computations than training, however, the inference is performed multiple times.
Hence, reducing the amount of computation during the inference is an
interesting ongoing goal~\cite{han2016eie}. Moreover, modern DNNs usually apply
the same number of operations for every inputs, and the natural question that
arises is whether this amount of computation is indeed
required~\cite{panda2016conditional}.

In this paper, we focus on the computational effort spent on inference in DNNs.
For simplicity, we measure the computational effort in the number of
multiply-accumulate operations (MACs).  Many claim that the computational
effort required for classifying images should depend on the
image~\cite{graves2016adaptive,figurnov2017spatially,panda2016conditional,teerapittayanon2016branchynet}.
We claim that the required computational effort for classification is an
intrinsic yet hidden property of the inputs. Namely, some images are much
easier to classify than others, but the required computational effort needed
for classification is hard to predict before classification is completed.

The desire to spend the ``right'' computational effort in classification leads
to the first goal in this work.
\begin{goal}
  Given a model $M$, design a model $M'$ in which the computational effort
  during the classification of an input $x$ is proportional to the likelihood of
  misclassifying $x$ using $M$.
\end{goal}

Misclassification likelihood indicates and measures the hardness of an input.
The question we pose is whether we can (almost) preserve accuracy while
reducing the computational effort required to classify "easy" instances. The
two extreme cases are: (1) Consider a distribution of inputs $D$ for which the
misclassification likelihood is very low (say $1\%$) in model $M$. We view $D$
as a distribution of "easy" inputs, and would like the new model  $M'$ to
classify $x\in D$ while spending a fraction of the computational effort
compared to $M$. (2) Consider a distribution of inputs $D'$ for which the
misclassification likelihood is high (say, $25\%$) in model $M$. We view $D'$
as a distribution of "hard" inputs, and would like the new model $M'$ to
classify inputs from $D'$  almost as accurately as $M$ does. The computational
effort of $M'$ for inputs in $D'$ is only slightly higher than that of $M$
(c.f., an overhead of $1\%$ in the computation). The principle behind our goal
is that an efficient model should achieve a high classification accuracy faster
for "easy" instances than for "harder" ones.

A motivation to reduce the computational effort during the inference can be
exemplified by systems with non-constant power consumption or throughput.
Examples of such settings are: (1)~As the battery drains in a mobile device,
one would like to enter a ``power saving mode'' in which less power is spent
per classification. (2)~If the input rate increases in a real-time system
(e.g., due to a burst of inputs), then one must spend less time per
input~\cite{cambazoglu2010early}. (3)~Timely processing in a data center during
spikes in query arrival rates may require reducing the computational effort per
query~\cite{bodik2010automating}.

Dynamic changes in the computational effort or the throughput lead to the
second goal in this work.
\begin{goal}
  Introduce the ability to dynamically control the computational
  effort while sacrificing accuracy as little as possible. Such changes
  in the computational effort should not involve retraining of the
  DNN.
\end{goal}

\subsection{Contribution}
We propose an architecture that is based on a cascade of
DNNs~\cite{bolukbasi2017adaptive} depicted in Figure~\ref{fig:flat}. The
cascade comprises multiple DNNs (e.g., three DNNs), called \emph{component
DNNs}. The cascade is organized sequentially so that the next component DNN is
fed by the previous component. Hence previous computations are reused and
further refined by the next component. Classification takes place by invoking
the component DNNs one-by-one and stopping the computation as soon as the
confidence level reaches the desired level. Our setting is applicable to
general multiclass classification in general architectures that terminate with
a softmax function.

The stopping decision is based on the softmax output of each component DNN.  We
define a simple confidence threshold, based on the softmax output, that allows
for trading off (a small) decrease in accuracy for (a substantial) reduction in
computational effort. The resulting approach has several advantages over the
previous
work~\cite{panda2016conditional,bolukbasi2017adaptive,stamoulis2018designing}.
The main contribution of our work is:

\medskip

\textbf{Dynamically change the compromise between accuracy and computational
effort without retraining the cascaded model.}

\medskip

In addition, we show how a cascaded architecture can be obtained from an
ordinary feed-forward DNN while requiring only small fine-tuning (see
section~\ref{sec:experiments}). We demonstrate the performance of our models on
various image classification datasets:     \begin{inparaenum}[(i)]
    \item A computation reduction of $34\%$ that sacrifices $1.2\%$
        accuracy with respect to the \cft test set.
    \item A computation reduction of $16\%$ that sacrifices $0.7\%$
        accuracy with respect to the \cfoh test set.
    \item A computation reduction of $54\%$ that sacrifices $1.4\%$
        accuracy with respect to the \svhn test set.
    \item A computation reduction of $17\%$ that sacrifices $1.3\%$
        accuracy with respect to the \inet validation set.
    \end{inparaenum}

Finally, our experimentation demonstrates a monotone relation between softmax
values and classification accuracy in intermediate classifiers (see
section~\ref{sec:softmax_linear_response}).

\section{Related work}\label{sec:related_work}
The two principle techniques that we employ are \emph{cascaded
  classification} and \emph{confidence estimation}. We elaborate on the recent usage of these techniques hereinafter.

\subsection{Cascaded classification}
Cascaded classification is suggested in the seminal work of Viola and
Jones~\cite{viola2001rapid}. As opposed to voting or stacking ensembles in
which classification is derived from the outputs of multiple experts (e.g.,
majority), the decision in a cascaded architecture is based on the last expert.
A cascaded neural network architecture for computer vision is presented
in~\cite{huang2018multi}. In their work, as the complexity of the input
increases, the evaluation is performed with increased resolution and increased
number of component DNNs in the cascade. The works of Wang
\etal~\cite{wang2017skipnet,wang2018sgad} presents the skipping approach, where
each input can take a path composed of a subset of layers of the original
architecture. Skipping of layers requires training of switches that decide
whether skipping of layers takes place. The work by Lerox
\etal~\cite{lerox2017cascading} presented the idea of early stopping in a
setting in which the cascaded DNNs are distributed among multiple devices.

Reinforcement learning is employed by Odena \etal~\cite{odena2017changing} in a
cascade of meta-layers to train controllers that select computational modules
per meta-layer. meta-layers to train controllers that select computational
modules per meta-layer.

\subsection{Confidence estimation}

Uncertainty measures of classifiers are discussed
in~\cite{cordella1995method,stefano2000reject}. These works address the issue
of the degree of confidence that a classifier has about its output. The
confidence of an assembly of algorithms is investigated by Fagin
\etal~\cite{fagin2003optimal} in general setup. Fagin \etal define instance
optimality and suggest to terminate the execution according to a criterion
based on a threshold.

Rejection refers to the event that a classifier is not confident about its
outcome, and hence, the output is rendered unreliable. Geifman and
El-Yaniv~\cite{geifman2017selective} describe a selective classification
technique, in which a classifier and a rejection-function are trained together.
The goal is to obtain coverage (i.e., at least one classifier does not reject)
while controlling the risk via rejection functions. They proposed a
softmax-response mechanism for deriving the rejection function and discussed
how the true-risk of a classifier (i.e., the average loss of all the
non-rejected samples) can be traded-off with its coverage (i.e., the mass of
the non-rejected region in the input space). Our work adopts the usage of the
softmax response as a confidence rate function, however, it differs in a way we
apply the confidence threshold. Namely, we propose a cascade of classifiers
that terminates as soon as the desired confidence threshold is reached.

The ability of the softmax output to reflect the true confidence of the
classifier was investigated by Gu \etal\cite{guo2017calibration}. The authors
propose the temperature scaling technique in order to calibrate the softmax
output, making it highly correlated with the expected accuracy.

\subsection{Combined approach: cascaded inference with confidence estimation}
The work of Cambazoglu \etal~\cite{cambazoglu2010early} presents an additive
ensemble machine learning approach with early exits in a context of the web
document ranking. In the additive approach, the sum of the outputs of a prefix
of the classifiers provides the current output confidence.

The work of Teerapittayanon \etal~\cite{teerapittayanon2016branchynet} presents
the BranchyNet approach, in which a neural network architecture has multiple
branches, each branch consists of a few convolutional layers terminated by a
classifier and a softmax function. The approach
in~\cite{teerapittayanon2016branchynet} does not help to reduce the amount of
computation that takes place outside the ``main path''. The confidence of an
output vector $y$ in BranchyNet is derived from the entropy function
$entropy(y)=-\sum_c y_c\log y_c$. Finally,
in~\cite{teerapittayanon2016branchynet}, automatic setting of threshold levels
is not developed, and the gains of their approach were not examined on large
datasets.

Cascaded classification with dedicated linear confidence estimations (rather
than softmax) appears in the Conditional Deep Learning (CDL)
of~\cite{panda2016conditional}, however, this approach was not examined on
large datasets and did not discuss an automatic setting of confidence
thresholds. Cascaded classification with confidence estimation appears also in
the SACT mechanism~\cite{figurnov2017spatially}, an extension of the prior work
by Graves~\cite{graves2016adaptive} that deals with recurrent neural networks.
Confidence estimation is based on the summation of the halting scores.
Computation is terminated as soon the cumulative halting score reaches a
threshold.  An interesting aspect of SACT architecture is the feature of
spatial adaptivity. Namely, different computational efforts are spent on
different regions of the input image.

Recently, Bolukbasi \etal~\cite{bolukbasi2017adaptive} proposed an
adaptive-early-exit cascaded classification architecture.  The computation may
terminate after each convolutional layer. For every convolutional layer $k$, a
special decision function $\gamma_k$ is trained to whether an exit should be
chosen. One of the drawbacks of this approach is that the decision functions
must be re-trained per value of the acceptable accuracy degradation.

%%%%%%%%%%%%%%%%%%%%%%%%%%%%%%%%%%%%%%%%%%%%%%%%%%%%%%%%%%%%%%%%%%%%%%%%%%
%%%%%%%%%%%%%%%%%%%%%%%%       Architecture      %%%%%%%%%%%%%%%%%%%%%%%%%
%%%%%%%%%%%%%%%%%%%%%%%%%%%%%%%%%%%%%%%%%%%%%%%%%%%%%%%%%%%%%%%%%%%%%%%%%%

\section{Cascaded Inference (CI)}\label{sec:ci}
Table~\ref{tab:params} lists the parameters and notations introduced in this
chapter.
\begin{table}[htbp]
  \centering
  \caption{Notations and definitions used in this paper}
    \begin{tabular}{ccl}
      Notation\bigstrut[b] & Domain & Semantics \\
      \toprule
      $\numepochs$\bigstrut[t] & $\mathbb{N}$ & Number of training epochs \\
      $\nummodels$ & $\mathbb{N}$  & Number of component DNNs in the cascade\\
      $\numclasses$ & $\mathbb{N}$  & Number of classes in the classification task  \\
      $n$& $\mathbb{N}$   & Number of \rs-blocks in a \rs-module  \\
      $T$ &  & Labeled training set, containing pairs of inputs and corresponding labels \\
      $M$ &  & Set of component DNNs that form a cascade  ($|M|=\nummodels$) \\
      $M_m$ &  & The $m^{th}$ component in the cascade, $m\in\{0,...,\nummodels-1\}$ \\
      $\theta_{conv_m}$ &  & Weights and biases of the convolutional layers in component $M_m$\\
      $\Theta_{conv}$ &  & Weights and biases of the convolutional layers in the cascade \\
      $\Theta_{fc}$ &  & Weights and biases of the fully connected layers of the cascade \\
      $\theta_{fc_m}$ &  & Weights and biases of the fully connected layers of component $M_m$\\
      $out_m(x)$ & $\{0,...,\numclasses-1\}$ & Class predicted by component $M_m$  for input $x$\\
      $\delta_m(x)$ & $[0,1]$ & Confidence output by component $M_m$ for input $x$\\
      $\hat\delta_m$ & $[0,1]$ & Confidence threshold of component $m$\\
%      $\vec{\delta}$ & $[0,1]^{\nummodels}$ & Vector of classifier-specific confidence thresholds \\
%      $\Delta$  & $[0,1]^{\nummodels\times \numclasses}$ & Matrix of (classifier,class)-specific confidence thresholds\\

    \end{tabular}%
  \label{tab:params}%
\end{table}%

Note: throughout the paper, we use the terms ``classifier'' and ``component
DNN'' interchangeably.

\subsection{Cascaded architecture}
A cascade of DNNs is a chain of convolutional layers with branching between
layers to a classifier (see Figure~\ref{fig:flat}).  Early termination in
cascaded DNN components means that intermediate feature maps are evaluated by
classifiers. These classifiers attempt to classify the feature map and output a
confidence measurement of their classification. If the confidence level is
above a threshold, then execution terminates, and the classification of the
intermediate feature map is output. See Figure~\ref{fig:flat} for an example of
a cascaded architecture based on three convolutional layers. Each component in
a cascaded architecture consists of convolutional layers followed by a
branching that leads to (1)~a classifier, and (2)~the next component.

In our experimentation, we employ \rs block layers \cite{he2015deep} as
component DNNs in our cascade. Moreover, in section~\ref{sec:experiments} we
show how a large pre-trained model (\rsl) can be quickly transformed into a
cascaded architecture.

\begin{figure}[ht!]
\centering
   \input{Flat_and_Gradual}
   \caption{An example of a cascaded architecture of three component DNNs with
     early termination. A cascade of convolutional layers
     $(CONV_0,\ldots,CONV_2)$ ends with a classifier \clf{2}. Early termination
     is enabled by introducing the classifiers \clf{i} after convolutional
     layers. Each classifier outputs a classification $out_i$
     and a its confidence $\delta_i$.
}
   \label{fig:flat}%
\end{figure}

It is tempting to adjust the aforementioned topology of the cascade for even
higher computational reuse. For example, the $out_i$ output of the classifier
can be fed to the following component (namely to the $CONV_{i+1}$). Such an
adjustment, however, is not applicable in the case when the $CONV$ layers are
pre-trained in a non-cascaded setup, losing a major advantage of the method we
propose.

\subsection{Early termination based on confidence threshold}
The usage of the threshold for determining early termination in the cascade is
listed as Algorithm~\ref{alg:hierarchical}.  The algorithm applies the
component DNNs one by one and stops as soon as the confidence measure reaches
the confidence threshold of this component. This approach differs from previous
cascaded architectures in which a combination (e.g., sum) of the confidence
measures of the components is used to control the
execution~\cite{figurnov2017spatially,cambazoglu2010early}.

Instead of having $\nummodels$ per-component thresholds, one could suggest
using a single global threshold for the whole cascade. Another alternative is
to set $\nummodels\cdot\numclasses$ thresholds for every (component, class)
pair. We empirically compared the three aforementioned approaches and found the
first one (per component thresholds) to be the most effective, which therefore
became the approach of our choice.

\begin{algorithm}[ht!]
  \caption[Early Termination]{\ci- Cascaded Inference. Early
    termination takes place as soon as the confidence level reaches
    the confidence threshold. }
    \begin{algorithmic}[1]
    \STATE {\bfseries Input:} cascaded model $M$, thresholds $\hat\delta$, input $x$
    \FOR{$m=0$ \TO $\nummodels-1$}
        \STATE $(out_m(x),\delta_m(x) )\leftarrow M_m (x)$
        \IF{$\delta_m(x)\ge\hat\delta_m$}
            \RETURN $out_m(x)$
        \ENDIF
    \ENDFOR
    \RETURN $out_{\nummodels-1}(x)$
    \end{algorithmic}
    \label{alg:hierarchical}
    \end{algorithm}

\subsection{Softmax confidence}
Every component DNN is terminated by a classifier with one or more FC layers
followed by a softmax function.  Let $z_m\in\mathbb{R}^{\numclasses}$ denote
the input to the softmax function in the $m$'th component of the cascade. Let
$s_m\in[0,1]^{\numclasses}$ denote the softmax vector in the $m$'th component.
The softmax vector is defined as follows.
\begin{definition}[softmax]\label{def:softmax} $s_m[i] =
%softmax_i(z_m)=
\frac{e^{z_m[i]}}{\sum_{c=0}^{\numclasses-1}e^{z_m[c]}}$.
\end{definition}
\begin{definition}[confidence measure]\label{def:delta}
  The confidence measure $\delta_m\in[0,1]$ is defined by
  $\delta_m \triangleq \max_c\{s_m[c] \mid 0\le c\le\numclasses-1\}$.
\end{definition}
\begin{definition}[predicted class]\label{def:out}
  The predicted class $out_m\in\{0,\ldots,\numclasses-1\}$ is defined to be the
  class $c$ such that $s_m[c]=\delta_m$.
\end{definition}

\section{Training procedure}\label{sec:training}
In this section we present the training procedure of the component DNNs.

Consider a cascaded architecture with $\nummodels$ components. We denote this
cascade by $M=(M_0,\ldots,M_{\nummodels-1})$, where $M_m$ denotes the $m$'th
component in the cascade.
Let $\Theta_{conv}=\{\theta_{conv_0},\ldots,\theta_{conv_{\nummodels-1}}\}$
denote the weights and biases of the convolutional layers of the component DNNs
$(M_1,\ldots,M_{\nummodels-1})$.
Let $\Theta_{clf}=\{\theta_{clf_0},\ldots,\theta_{clf_{\nummodels-1}}\}$ denote
the weights and biases of the classifiers of the component DNNs
$(M_1,\ldots,M_{\nummodels-1})$.

Let $L_M(out_m,T)$ denote a loss function of the cascade $M$ with respect to
the output of the $m$'th component, averaged over the labeled dataset $T$. In
order to train the cascade $M$, we propose a backtrack-training
(Algorithm~\ref{alg:backtrack}) \bt. We emphasize that the training procedure
first optimizes all the convolutional weights together with the weights of the
last classifier. Only then, do we optimize the weights of the classifiers
\clf{i}, for $0\leq i \leq \nummodels-2$ (i.e., classifiers of intermediate
components). Our approach differs from previous training
procedures~\cite{teerapittayanon2016branchynet,wang2017skipnet} in which the
loss functions associated with all the classifiers were \emph{jointly
optimized}. This difference has two following advantages: (1) the longest
computational path of the cascade is trained independently of the intermediate
loss functions, hence the maximum achievable accuracy of the model is not
compromised. (2) A  pre-trained, non-cascaded, architecture can be transformed
into a cascade and then trained according to
lines~\ref{line:start_backtrack}-\ref{line:end_backtrack} of the \bt\ algorithm
to fine-tune only the intermediate classifiers.

    \begin{algorithm}[ht!]
    \caption [Cascaded Inference Backtrack training] {\bt\ - An algorithm for
    a backtrack training of the cascade
    $M=\{M_0,\ldots,M_{\nummodels-1}\}$. The output is the trained
    weights of the cascade $M$}
    \begin{algorithmic}[1]
        \STATE {\bfseries Input:} cascaded model $M$, training set $T$
        \STATE Let $\Theta_{deep} = \Theta_{conv}\cup
             \theta_{clf_{\nummodels-1}}$
        \STATE $\Theta_{deep}=\argmin_{\Theta_{deep}}\{L_M(out_{\nummodels-1},T)\}$.
        \FOR{$m=0$ \TO $\nummodels-2$}\label{line:start_backtrack}
            \STATE $\theta_{clf_m}=\argmin_{\theta_{clf_m}}\{L_M(out_m,T)\}$.
        \ENDFOR
        \RETURN $\Theta_{conv}\cup\Theta_{clf}$\label{line:end_backtrack}
    \end{algorithmic}
    \label{alg:backtrack}
    \end{algorithm}

\section{Setting of confidence thresholds}\label{sec:confidence_threshold_approach}
In this section, we present an automatic methodology for setting the confidence
threshold $\hat\delta_m$ for every component $M_m$ given an acceptable accuracy
degradation $\epsilon$. We note that the hyper-parameter $\epsilon$ is a single
parameter for the whole cascade, and the automatic methodology we present
determines an individual confidence threshold for every component in the
cascade. The important attribute of the automatic setting of the confidence
thresholds is that one can change them on the fly during the inference stage.

Let $T_m(\delta)\subseteq T$ denote the subset of inputs for which the
confidence measure of the $m$th component is at least $\delta$.

\begin{align*}
 T_m(\delta) &\triangleq \{(x,y)\mid \delta_m(x) \ge \delta\}.
\end{align*}

     Let $\gamma_m(\delta)$ denote the number of times the
   classification output by component $M_m$ is correct for inputs in
   $T_m(\delta)$.
   \begin{align*}
 \gamma_m(\delta) &\triangleq \sum_{(x,y)\in T_m(\delta)} \indicator{out_m (x) =
                y}.
   \end{align*}

         Let $\alpha_m(\delta)$ denote the accuracy of component $M_m$
         with respect to $T_m(\delta)$.
         \begin{align*}
 \alpha_{m}(\delta) &\triangleq \begin{cases}
                          \frac{\gamma_{m}(\delta)}{|T_{m}(\delta)|} \text{ if }|T_{m}(\delta)| > 0 \\
                          0 \text{ otherwise } \\
                          \end{cases}
         \end{align*}

Let $\alpha^*_m$ denote the maximum accuracy for component $M_m$.

\begin{align*}
  \alpha^*_m&\triangleq \max_{\delta\in [0,1]} \alpha_m(\delta).
\end{align*}

For an acceptable accuracy degradation $\epsilon>0$, we define the confidence
threshold $\delta_m(\epsilon)$ by
\begin{align*}
  \delta_m(\epsilon) \triangleq \min{}\{\delta \mid \alpha_m(\delta)\geq \alpha^*_m -\epsilon\}.
\end{align*}

When a cascaded inference is performed using \ci
(Algorithm~\ref{alg:hierarchical}), the confidence threshold vector
$\hat\delta$ is set as follows. Choose an $\epsilon\in[0,1]$, and set
$\hat\delta_m\gets \delta_m(\epsilon)$, for every $m$. We remark that (i) the
threshold for the last component should be zero, and (ii) one could use
separate datasets for training the weights and setting the confidence
threshold.

In some applications, the desired accuracy metric is ``top-$K$'', meaning that
a prediction is regarded as correct if the top $K$ most confident predictions
of a classifier contain the ground truth class. To choose appropriate
thresholds for the top-$K$ metric, the only change to the methodology above is
to set $\delta_m(x)$ to be the sum over the top $K$ elements in the softmax
vector.

%%%%%%%%%%%%%%%%%%%%%%%%%%%%%%%%%%%%%%%%%%%%%%%%%%%%%%%%%%%%%%%%%%%%%%%%%%
%%%%%%%%%%%%%%%%%%%%%%%%       Experiments       %%%%%%%%%%%%%%%%%%%%%%%%%
%%%%%%%%%%%%%%%%%%%%%%%%%%%%%%%%%%%%%%%%%%%%%%%%%%%%%%%%%%%%%%%%%%%%%%%%%%

\section{Experimental Setup}\label{sec:experiments}

In order to examine the usefulness of cascaded inference we performed
experiments on \cft, \cfoh\ and \svhn\ datasets using \rss architecture and on
\inet dataset using \rsl architecture. The transformation of the \rs
architecture into a cascade was done by dividing it into three stages; the
choice of the layers after which a new stage begins was based on the structure
of the architecture (i.e., between distinct layer blocks, differently
color-coded in Figure 3 of ~\cite{he2015deep}). In the resulting cascade, each
component DNN consists of a stage and a classifier, as depicted in
Figure~\ref{subfig:Resnet_arch_gradual_enhanced}. The analysis of the overhead
introduced by our transformation is depicted in Table~\ref{tab:mac}. According
to this analysis, the increase in the number of MACs, caused by the
transformation of the \rs\ into a cascade of $3$ component DNNs, is less than
$0.2\%$.

\begin{figure}[ht!]
    \centering
           \Mysubfloat{Building-Block}{0.19}{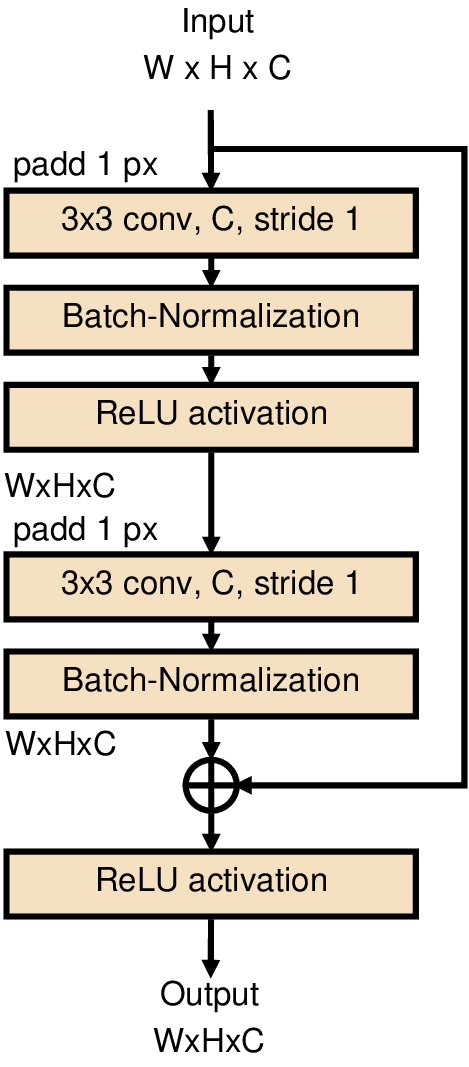}
                    {subfig:Resnet_block}
                    \hfill
           \Mysubfloat{Building-Block-i}{0.28} {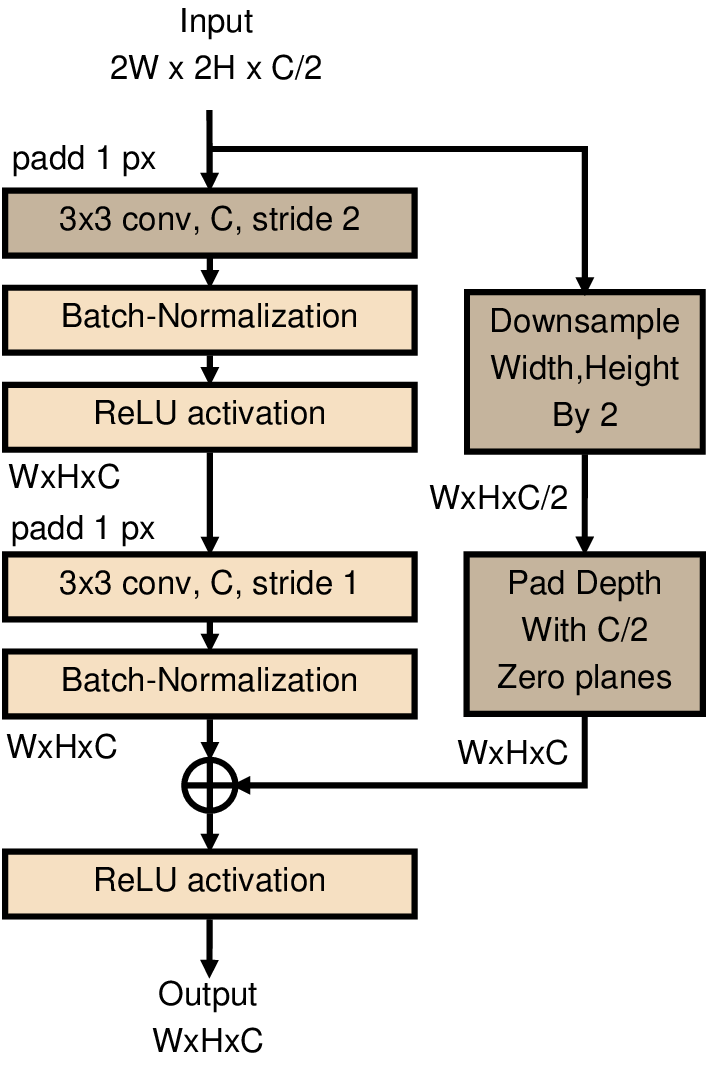}
                    {subfig:Resnet_block_first}
                    \hfill
           \Mysubfloat{Cascaded \rs}{0.32}{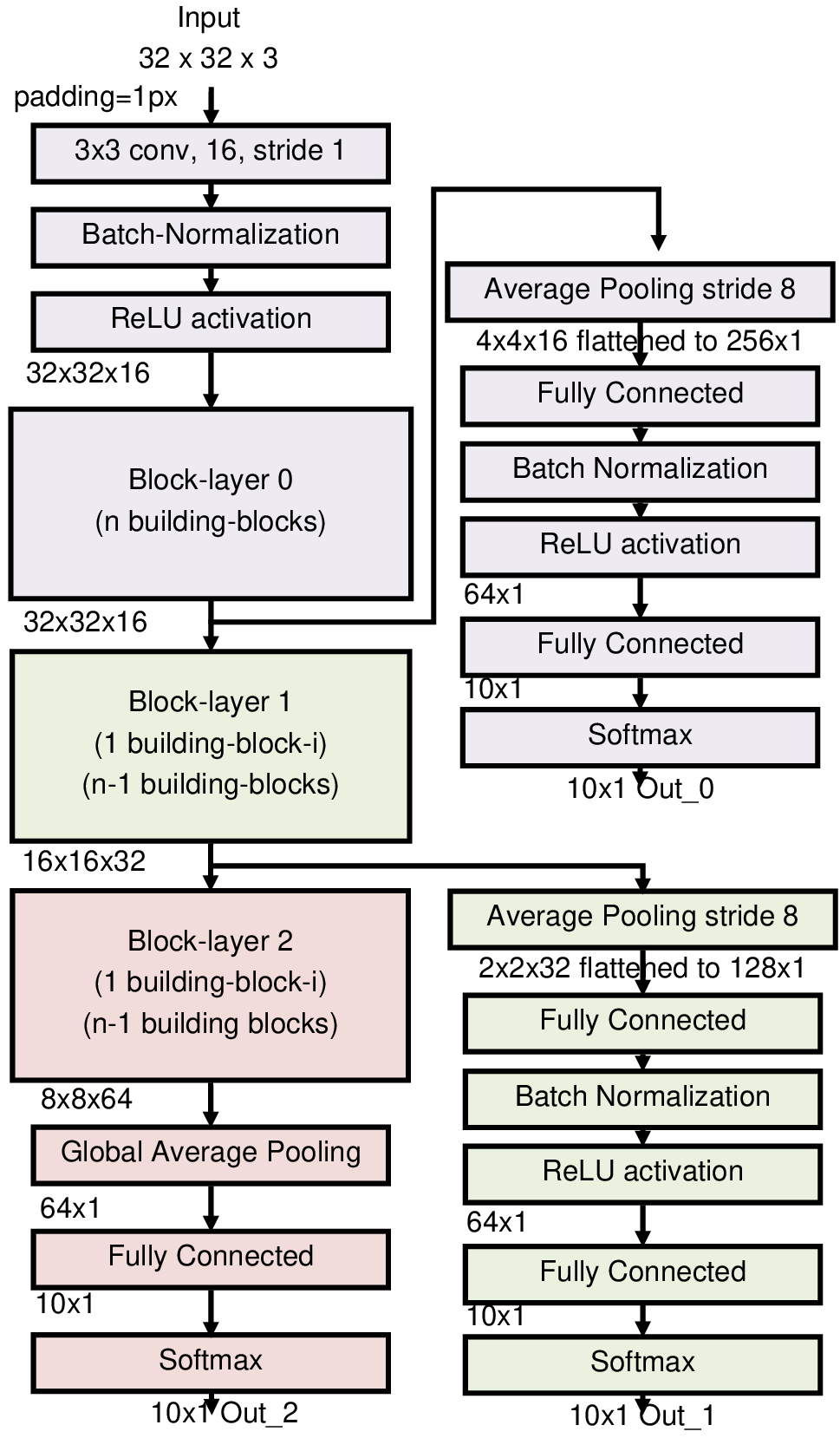}
                    {subfig:Resnet_arch_gradual_enhanced}
                    \hfill

       \caption{%\ref{subfig:Resnet_block}%
       (a) - building-block. %
       %\ref{subfig:Resnet_block_first}%
       (b) - first building-block in each block-layer performs a sub-sampling using stride 2. %
       %\ref{subfig:Resnet_arch_gradual_enhanced}%
       (c) - the cascaded version of \rs architecture, parameterized by $n$
       such that the number of layers in it is $2+6n$. For instance, setting
       $n=18$ yields the structure denoted \rs110 in the literature.}
       \label{fig:resnet_arch}%
    \end{figure}

\begin{table}[htbp]
   \centering
   \caption{ Number of MAC operations required for a single inference in ordinary \rs\ models and in their cascaded counterparts.}
 % Table generated by Excel2LaTeX from sheet 'Architecture Summary'
 \begin{tabular}{l|r|r}
       & \multicolumn{1}{l|}{ \rss} & \multicolumn{1}{l}{ \rsl} \\
 \midrule
  non-Cascaded &  $253,953,214$ &  $4,037,883,817$ \\
 \midrule
  Cascade - total &  $253,978,670$ &  $4,044,633,979$ \\
  component DNN 0 &  $86,000,922$ &  $1,817,092,009$ \\
  component DNN 1 &  $84,068,170$ &  $1,467,571,177$ \\
  component DNN 2 &  $83,909,578$ &  $759,970,793$ \\
 \midrule
  Computation increase &  $0.01\%$ &  $0.17\%$ \\
 \end{tabular}%

   \label{tab:mac}%
 \end{table}%

We trained the cascaded \rss from scratch with respect to
algorithm~\ref{alg:backtrack}. Simple data augmentation was employed only for
\cf\ models as in~\cite{he2015deep}. The optimization of every classifier was
performed with Stochastic Gradient Descent (SGD) for $160$ epochs in CIFAR
datasets and for $50$ epochs for \svhn dataset. Learning rate was scheduled as
described in~\cite{he2015deep}. For the \inet dataset we chose the
\rsl\cite{identity2016he} architecture, which requires roughly $4$ Giga-MAC
operations per inference and achieves a classification accuracy of roughly
$76.51\%$\footnote{\url{https://github.com/tensorflow/models/tree/master/official/resnet}\label{note1}}.
Deeper models such as \rs-$152$ and \rs-$200$ require more computation. One
could argue that these deeper architectures contain a great deal of redundant 
computation without sacrificing accuracy.

For the \inet experiments, we transformed the official pre-trained
Tensorflow~\cite{tensorflow2015whitepaper} \rsl model into a 3-stage cascaded
architecture by introducing additional classifiers after every layer block. We
then followed lines~\ref{line:start_backtrack}-\ref{line:end_backtrack} of the
\bt\ algorithm (Algorithm \ref{alg:backtrack}) to train the two new
classifiers, each of which has $2$ FC layers, while freezing all the
pre-trained weights. Source code, for reproducing our \inet results, is
publicly
available\footnote{\url{https://github.com/AnonymousConferenceCode/Cascaded_Inference}.}.
This fine tuning of the pre-trained model took less than $20$ hours per
classifier using $4$ GPUs.

\section{Results}\label{sec:results}
\subsection{Confidence threshold effect}

\begin{figure}[htbp]
   \centering
   \Mysubfloat{\cft}{0.499}
            {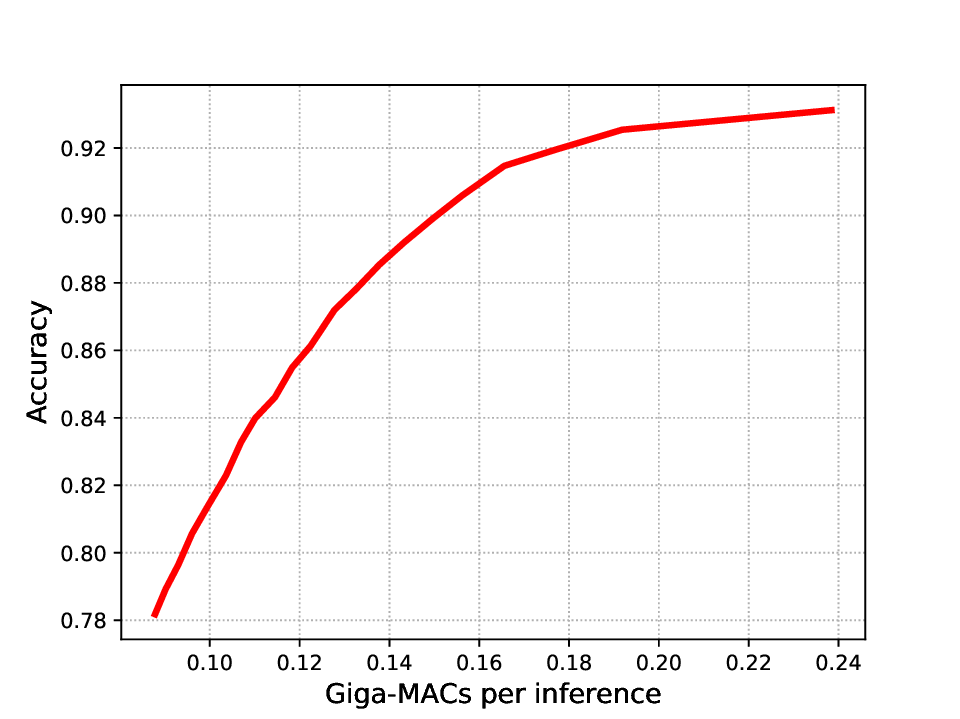}
            {subfig:concluding_plot_cft}
   \Mysubfloat{\cfoh}{0.499}
            {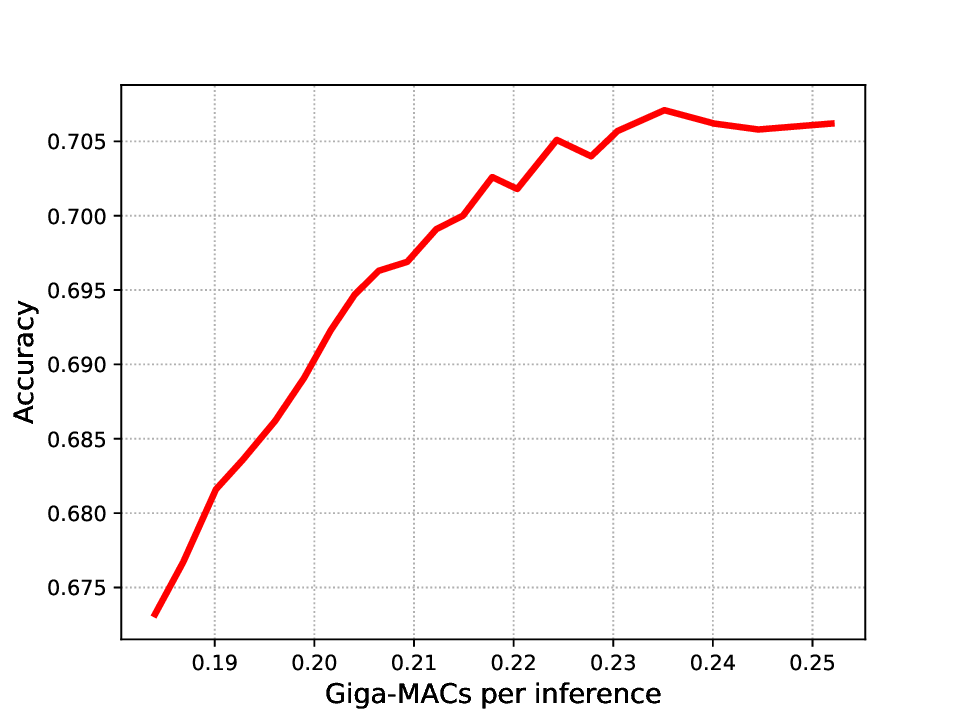}
            {subfig:concluding_plot_cfoh}

   \Mysubfloat{\svhn}{0.499} {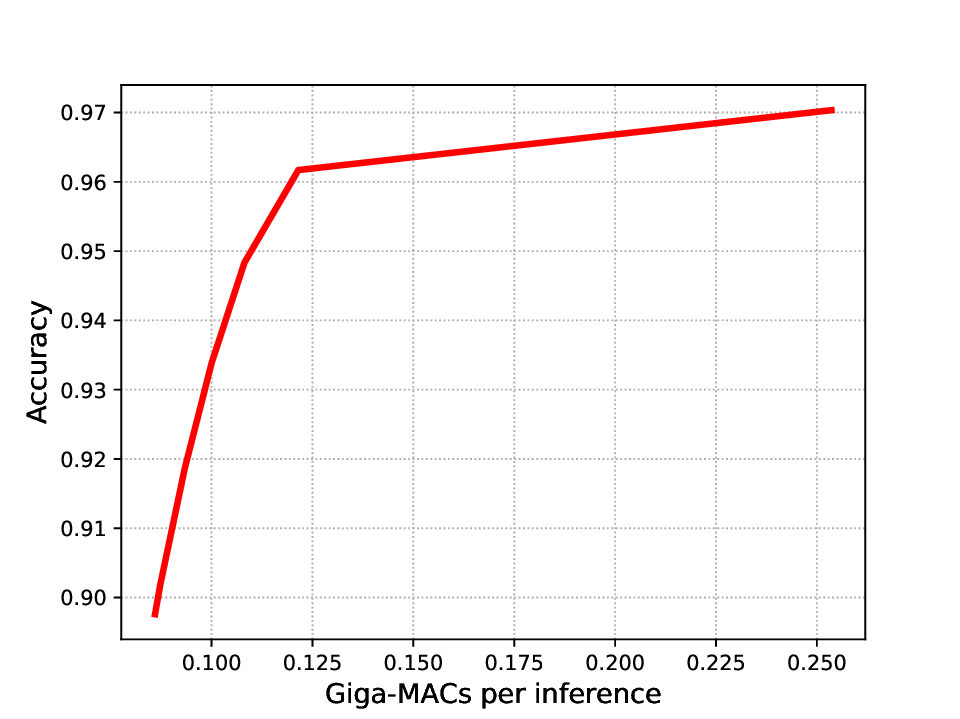}
            {subfig:concluding_plot_svhn}%
   \Mysubfloat{\inet top-1}{0.499} {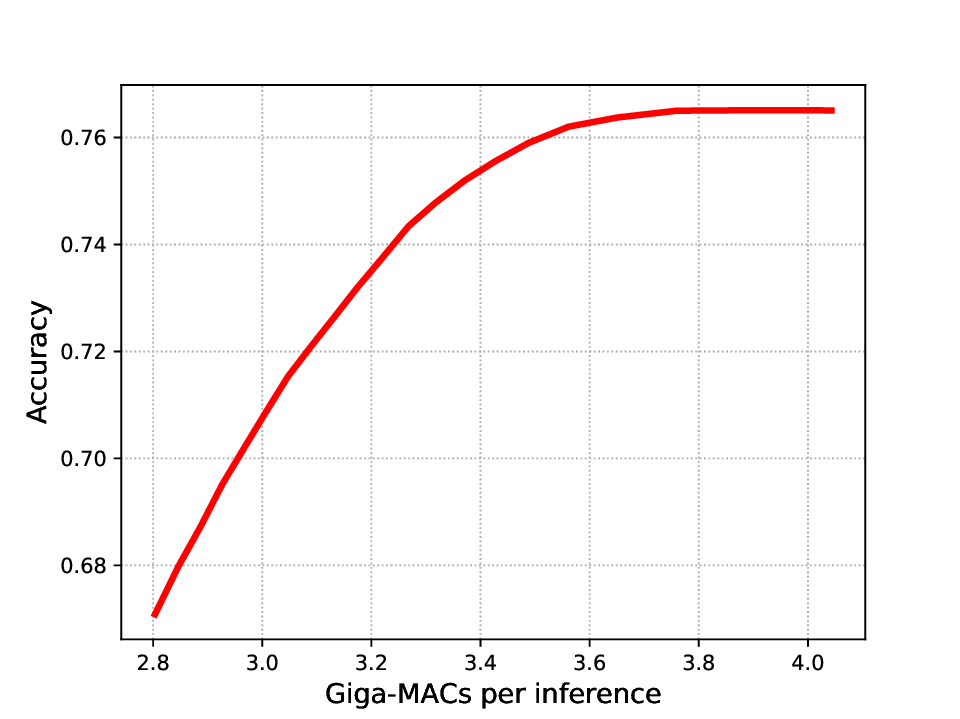}
            {subfig:concluding_plot_imagenet_top1}%

   \Mysubfloat{\inet top-5}{0.499} {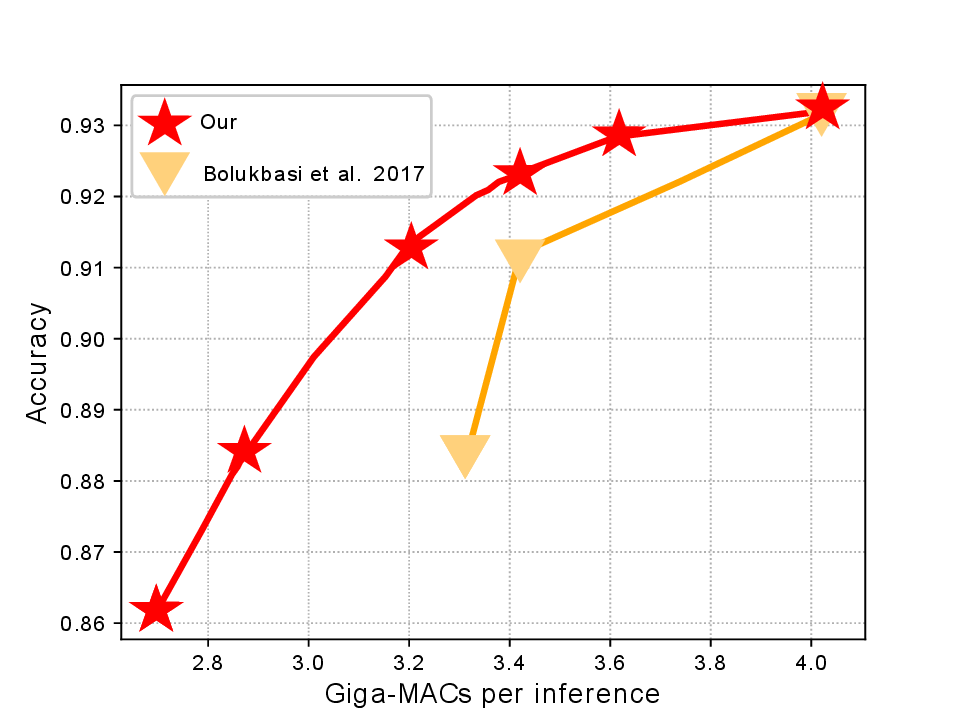}
            {subfig:concluding_plot_imagenet_top5}%
            \caption{\textbf{Cascaded inference with early
                termination} test accuracy vs. average number of
              MAC operations per inference. The measured points on
              the curves are obtained by considering variable
              values of $\epsilon\in\{20\%,\ldots,1\%,0\%\}$.}
   \label{fig:cst_acc_mac}%
\end{figure}

We trained the cascaded versions of \rss and \rsl models as described in
Section~\ref{sec:experiments}. We evaluated the performance using various
$\epsilon$ values. The tradeoff between the test-accuracy and the number of
MACs required for a single inference is shown in Figure~\ref{fig:cst_acc_mac}.
The MAC counts were obtained analytically by summing up the linear operations
in the convolutional layers and the FC layers, excluding activations and batch
normalization. Quantitative results that appear in Table~\ref{tab:performance}
demonstrate the ability of the cascaded architectures to trade as little as
$1.3\%$ of accuracy for a reduction of $16\%-53\%$ of the computational effort.
Note the reduced effect on accuracy for \inet when accuracy is measured with
respect to the top-$5$ classifications compared to the top-$1$ classification
(see last two lines in Table~\ref{tab:performance}).

\begin{table*}[htbp]
{\scriptsize
\centering
\caption{\textbf{Accuracy-computation tradeoffs.} - Column $1$ lists
  the tested datasets.  Columns $2$-$4$ list the accuracy of
  classifier \clf{i}, for $i\in\{0,1,2\}$, with respect to the
  complete test set.  Columns $5$-$10$ list the accuracy of our
  cascaded architecture for different values of $\epsilon$ - the
  acceptable accuracy degradation (see Sec.~\ref{sec:experiments} for
  the details of which network was used for each
  dataset). Computational reduction by the cascade for each $\epsilon$
  is relative to the computational effort of the non-cascaded
  architecture $M_{0,1,2}$ and is defined by
  $1-\frac{\#MAC\_Count(Cascade(\epsilon))}{\#MAC\_Count(M_{0,1,2})}$.}

    % Table generated by Excel2LaTeX from sheet 'Performance'
     \begin{tabular}{lccccccccc}
       & \multicolumn{3}{c}{Accuracy of $M_{0,\ldots,m-1}$} & \multicolumn{6}{c}{Accuracy(top), computation reduction(bottom)} \\
       \midrule
       \multicolumn{1}{c}{Dataset} & $M_0$   & $M_{0,1}$   & $M_{0,1,2}$   & $\epsilon=0\%$ & $\epsilon=1\%$ & $\epsilon=2\%$ & $\epsilon=4\%$ & $\epsilon=7\%$ & $\epsilon=8\%$ \\
       \midrule
       \multirow{2}[2]{*}{\cft} & \multirow{2}[2]{*}{$77.50\%$} & \multirow{2}[2]{*}{$81.40\%$} & \multirow{2}[2]{*}{$93.10\%$} & $93.10\%$ & $92.70\%$ & $\bm{91.90\%}$ & $91.10\%$ & $87.32\%$ & $86.35\%$ \\
       &       &       &       & $6\%$ & $27\%$ & $\bm{34\%}$ & 42\% & $50\%$ & $52\%$ \\
       \midrule
       \multirow{2}[2]{*}{\cfoh} & \multirow{2}[2]{*}{48.10\%} & \multirow{2}[2]{*}{50.00\%} & \multirow{2}[2]{*}{70.50\%} & 70.50\% & 70.65\% & $70.50\%$ & $70.30\%$ & $69.94\%$ & $\bm{69.78\%}$ \\
       &       &       &       & $1\%$ & $4\%$ & $7\%$ & $10\%$ & $15\%$ & $\bm{16\%}$ \\
       \midrule
       \multirow{2}[2]{*}{\svhn} & \multirow{2}[2]{*}{$89.80\%$} & \multirow{2}[2]{*}{$85.20\%$} & \multirow{2}[2]{*}{$97.03\%$} & $97.03\%$ & $\bm{95.60\%}$ & $94.00\%$ & $91.30\%$ & $89.76\%$ & $89.80\%$ \\
       &       &       &       & $0\%$ & $\bm{54\%}$ & $59\%$ & $64\%$ & $66\%$ & $66\%$ \\
       \midrule

         \multicolumn{1}{c}{ } &     &     &     & $\epsilon=0\%$ & $\epsilon=1\%$ & $\epsilon=2\%$ & $\epsilon=5\%$ & $\epsilon=6\%$ & $\epsilon=7\%$ \\
         \midrule
         \inet & \multirow{2}[2]{*}{46.69\%} & \multirow{2}[2]{*}{$62.76\%$} & \multirow{2}[2]{*}{$76.51\%$} & $76.51\%$ & $76.51\%$& $76.50\%$ & $75.90\%$ & $75.56\%$ & $\bm{75.19\%}$ \\
            top-1   &       &       &       & $0\%$   & $3\%$   & $7\%$   & $14\%$  & $15\%$  & $\bm{17\%}$ \\
         \midrule
         \inet & \multirow{2}[2]{*}{$70.22\%$} & \multirow{2}[2]{*}{$84.31\%$} & \multirow{2}[2]{*}{$93.21\%$} & $93.21\%$ & $92.84\%$ & $\bm{92.02\%}$ & $88.54\%$ & $87.30\%$ & $86.13\%$ \\
            top-5   &       &       &       & $0\%$   & $11\%$  & $\bm{17\%}$  & $28\%$  & $31\%$ & $33\%$\\
         \bottomrule
         \end{tabular}%

    \label{tab:performance}%
    }
\end{table*}%

\subsection{Comparison with Bolukbasi \etal}
Figure~\ref{subfig:concluding_plot_imagenet_top5} and Table~\ref{tab:bolukbasi}
compare the top-$5$ accuracy-to-computation tradeoffs of our cascaded inference
against adaptive cascaded inference over \rs-$50$ with early
exits~\cite{bolukbasi2017adaptive}. We translated the speedups presented
in~\cite{bolukbasi2017adaptive} from time to MAC-count speedups for the purpose
of comparison to our work (to exclude the impact of software and hardware
environments differences). Our model demonstrates higher accuracy for any given
computational effort, in addition to being able to dynamically adjust to
different accuracy-to-computation tradeoffs.

% Table generated by Excel2LaTeX from sheet 'Performance'
\begin{table}[htbp]
  \centering \caption{\textbf{Comparison of cascaded inference to
      \cite{bolukbasi2017adaptive}} on \inet top-5 metric.  Column $1$ lists
    the accuracy lost relative to the original \rs-50 model. Columns
    $2$ and $3$ list the speedups of the work by Bolukbasi \etal\ and
    of our cascaded inference with respect to the full \rs-50
    model. Speedup is $\frac{old\_MACs}{new\_MACs}-1$.}
    \begin{tabular}{c|c|c}
    Accuracy reduction& Bolukbasi et al. 2017 speedup  & Our Cascade speedup  \bigstrut[b]\\
    \hline
    $1\%$   & $8\%$   & $20\%$\\% @ $\epsilon=1.8\%$  \bigstrut[t]\\
    $2\%$   & $18\%$  & $27\%$\\% @ $\epsilon=2.8\%$\\
    $5\%$   & $22\%$  & $41\%$% @ $\epsilon=5.3\%$ \\
    \end{tabular}%
  \label{tab:bolukbasi}%
\end{table}%

\subsection{Softmax as a confidence measure}\label{sec:softmax_linear_response}

\begin{figure}[htbp]
    \centering
    \Mysubfloat{\cft}{0.499}%
            {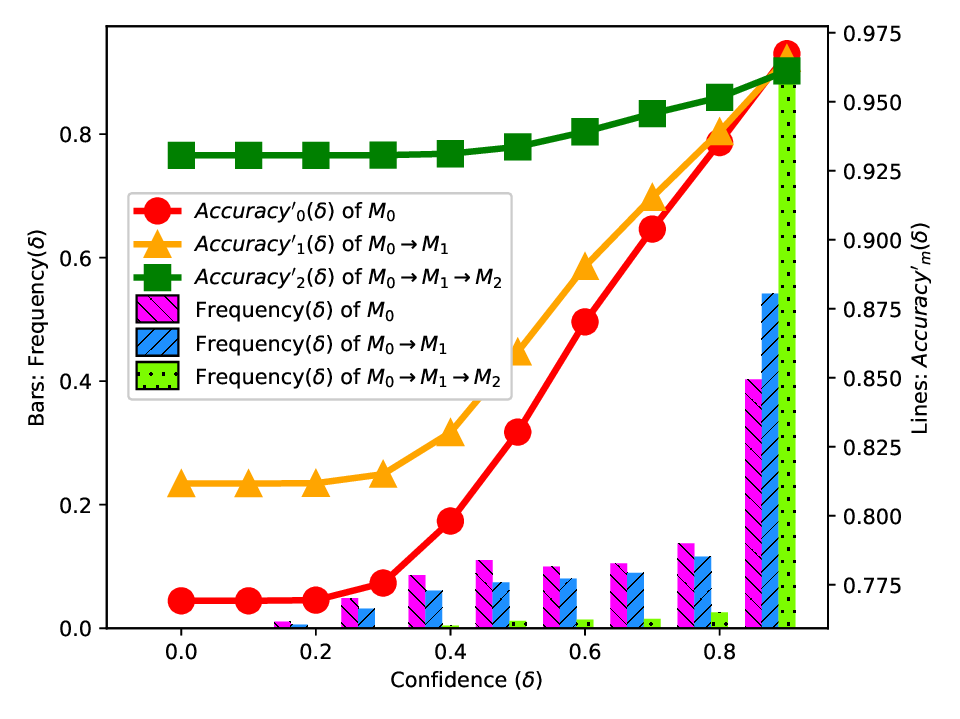}%
            {subfig:resnet_cft}
   \Mysubfloat{\cfoh}{0.499}%
            {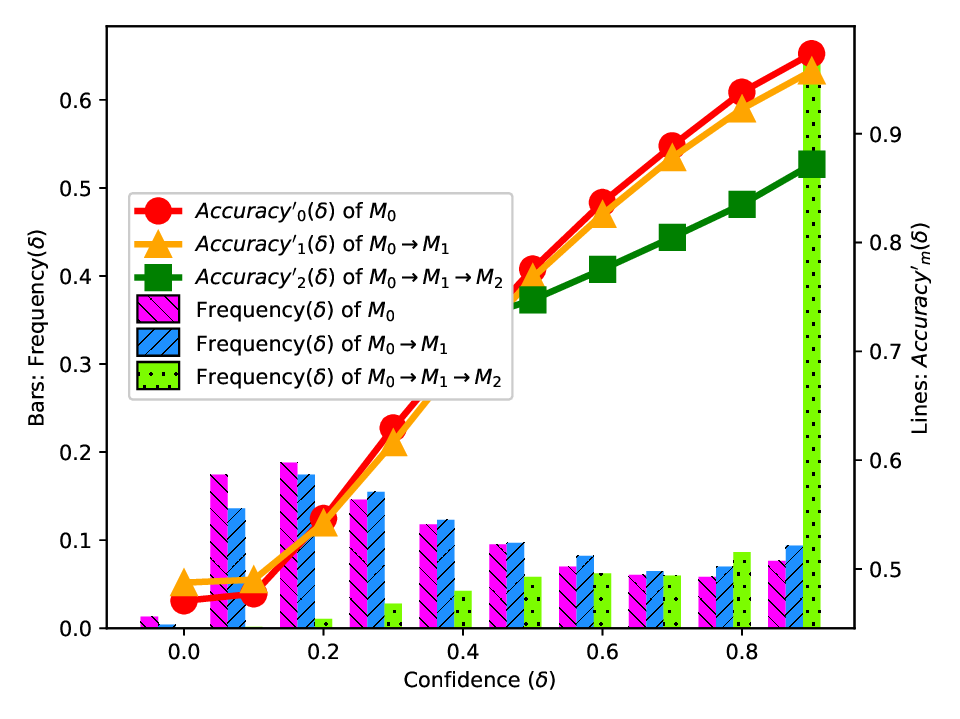}%
            {subfig:resnet_cfoh}

   \Mysubfloat{\svhn}{0.499}%
            {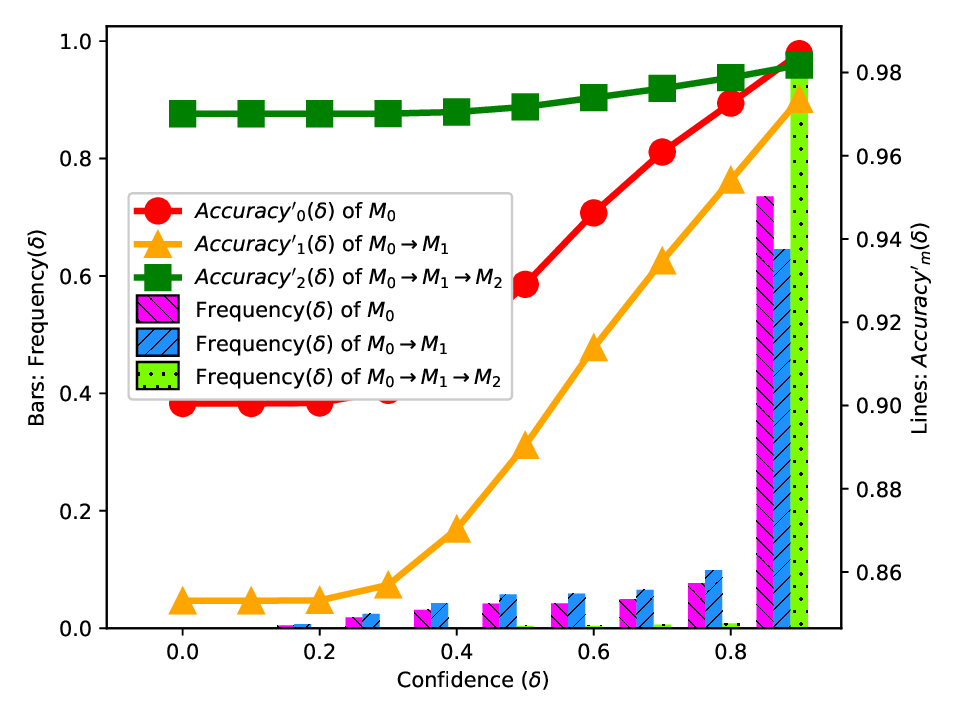}%
            {subfig:resnet_svhn}%
   \Mysubfloat{\inet}{0.499}%
            {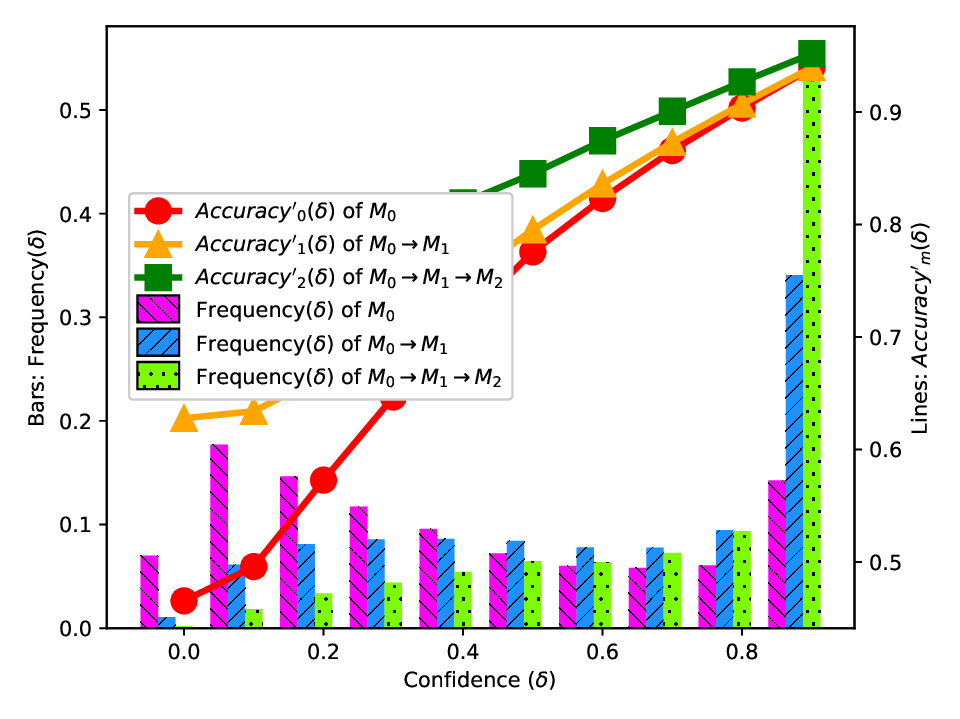}%
            {subfig:resnet_imagenet}%
            \caption{\textbf{Softmax as a confidence measure}. The line plots
            show the accuracy $\alpha_m(\delta)$ of each
            classifier in the cascade independently. The bar plot
            presents the frequency of the different confidence levels sampled
              over the test set. All plots were obtained by separately testing
              the three component DNNs of the cascaded \rs.}
    \label{fig:confidence_characteristics}
    \end{figure}

For the cascaded \rs models, we analyzed the accuracy $\alpha_m(\delta)$ (see
definition in Section~\ref{sec:confidence_threshold_approach}) of each
classifier independently. The accuracy $\alpha_m(\delta)$ was measured for
$\delta\in[0,1]$ using the test-set rather than to the training set. The plots
in Figure~\ref{fig:confidence_characteristics} show how the choice of the
threshold provides control over the test accuracy. Note that the range of
$\alpha_m(\delta)$ starts with the accuracy of \clf{m} and ends with the
accuracy that corresponds to the highest confidence measure.  The almost linear
behavior of $\alpha_m(\delta)$ as a function of $\delta$ justifies basing the
confidence threshold on the softmax output. We note that these results were
obtained without applying softmax calibration techniques.

In addition, we examined the frequency of the different confidence levels
observed at the output of each classifier in a cascade. This observation is
presented in the form of a bar-plot distribution in
Figure~\ref{fig:confidence_characteristics}. The distribution of the first two
components of the cascade is relatively uniform. Whereas the distribution of
the confidences of the last classifier has no importance since in our inference
approach the confidence threshold of the last classifier is set to
$\hat\delta_{\nummodels-1}=0$.

\section{Discussion and future work}\label{sec:discussion}

As further research, cascading can be applied to RNNs or alternatively, the
impact of depth of feed-forward DNNs on the confidence estimation can be
investigated. A gap between the allowed accuracy degradation ($\epsilon$) and
the actual test accuracy degradation was especially evident in the \cfoh
dataset. This gap can be bridged by performing softmax-calibration, which can
serve as a practical extension of our study.

%%%%%%%%%%%%%%%%%%%%%%%%%%%%%%%%%%%%%%%%%%%%%%%%%%%%%%%%%%%%%%%%%%%%%%%%%%
%%%%%%%%%%%%%%%%%%%%%%%%        Conclusion       %%%%%%%%%%%%%%%%%%%%%%%%%
%%%%%%%%%%%%%%%%%%%%%%%%%%%%%%%%%%%%%%%%%%%%%%%%%%%%%%%%%%%%%%%%%%%%%%%%%%

\section{Conclusions}\label{sec:conclusion}

We showed that using a softmax output as a confidence measure in a cascade of
DNNs can provide a reduction of $15\%-50\%$ in the number of MAC operations
while degrading the classification accuracy by roughly $1\%$. This approach
allows to dynamically change the acceptable accuracy degradation ($\epsilon$)
without retraining because the confidence thresholds are automatically derived
from $\epsilon$. This achieves the second goal of our work.

Secondly, our approach  is easily adoptable, since the transformation of the
trained non-cascaded DNN into a cascade of component DNNs requires only
training of the auxiliary classifiers, which are small relative to the original
network. In other words, non-cascaded state-of-the-art models can be
transformed into a cascade of component DNNs with very little training
involved. Once the transformation is complete, these models will benefit from
less computation during inference.

Finally, we observed a monotone, almost linear, relation between the softmax
function and the test accuracy. This implies that the softmax output is a good
estimate of the neural network confidence. Our approach explicitly demands
lower computational effort for inputs that indicate higher confidence. This
achieves the first goal of this work.

\subsubsection*{Acknowledgments}
We thank Nissim Halabi, Moni Shahar and Daniel Soudry for useful conversations.

\bibliographystyle{alpha}
\bibliography{references}

\end{document}